\documentclass{article} 
\pdfoutput=1
\usepackage[utf8]{inputenc}
\usepackage[english]{babel}
\usepackage[T1]{fontenc}
\usepackage{amsmath}
\usepackage{hyperref}
\usepackage{tikz}
\usepackage{lipsum}
\usepackage[letterpaper,inner=3.5cm,outer=2.5cm,top=2.5cm,bottom=2.5cm,pdftex]{geometry}
\usepackage[T1]{fontenc}
\usepackage[utf8]{inputenc}
\usepackage{authblk}
\usepackage{fancyhdr}
\usepackage{url}            
\usepackage{booktabs}       
\usepackage{amsfonts}       
\usepackage{nicefrac}       
\usepackage{microtype}      
\usepackage{subfigure}
\usepackage{braket}
\usepackage{amsthm}
\usepackage{amssymb}
\usepackage{mathtools}
\newtheorem{theorem}{Theorem}
\usepackage[square,comma,numbers,compress]{natbib}
\newcommand{\mute}[1]{}
\newcommand{\erf}{\mathrm{erf}\,}

\newcommand{\Rset}{\mathbb{R}}

\newcommand{\me}{e}

\newcommand{\argmin}{\operatornamewithlimits{arg\ min}}

\newcommand{\spc}{{\;\;\;}}
\newcommand{\half}{\frac{1}{2}}
\newcommand{\st}{\mathrm{s.t.}}

\newcommand{\BigTheta}[1]{\Theta\left({#1}\right)}
\newcommand{\BigOh}[1]{O\left({#1}\right)}
\newcommand{\BigPolyOh}[1]{\tilde{O}\left({#1}\right)}

\newcommand{\BigOmega}[1]{\Omega\left({#1}\right)}

\newcommand{\sstar}{{\star}}
\newcommand*\diff{\mathop{}\!\mathrm{d}}

\newcommand{\norm}[1]{||#1||}

\newtheorem{lemma}[theorem]{Lemma}
\newtheorem{mydef}[theorem]{Definition}

\newcommand{\majV}{f}

\newcommand{\floor}[1]{ \lceil #1 \rceil}

\title{Quantum Sparse Support Vector Machines}

\author{Seyran Saeedi\thanks{Current address: Department of Electrical and Computer Engineering, University of California, Santa Barbara, CA, USA. e-mail: seyran@ucsb.edu}}
\author{Tom Arodz\thanks{Corresponding author. e-mail: tarodz@vcu.edu}}
\affil{\mbox{Department of Computer Science}, \mbox{Virginia Commonwealth University} \mbox{Richmond, VA, USA}}
\date{}

\begin{document}
	\sloppy
	\maketitle

\begin{abstract}
We analyze the computational complexity of Quantum Sparse Support Vector Machine, a linear classifier that minimizes the hinge loss and the $L_1$ norm of the feature weights vector and relies on a quantum linear programming solver instead of a classical solver. Sparse SVM leads to sparse models that use only a small fraction of the input features in making decisions, and is especially useful when the total number of features, $p$, approaches or exceeds the number of training samples, $m$. We prove a $\Omega(m)$ worst-case lower bound for computational complexity of any quantum training algorithm relying on black-box access to training samples; quantum sparse SVM has at least linear worst-case complexity. However, we prove that there are realistic scenarios in which a sparse linear classifier is expected to have high accuracy, and can be trained in sublinear time in terms of both the number of training samples and the number of features.
\end{abstract}

\section{Introduction}
Steadily increasing ability to measure large number of features in large number of samples leads to ongoing interest in fast methods for solving large-scale classification problems. One of the approaches being explored is training the predictive model using a quantum algorithm that has access to the training set stored in quantum-accessible memory. In parallel to research on efficient architectures for quantum memory \cite{blencowe2010quantum,park2019circuit,jiang2019experimental}, work on quantum machine learning algorithms and on quantum learning theory is under way (see  \cite{BWPR17,dunjko2018machine,SchF18,arunachalam2017guest} and \cite{dunjko2020non} for reviews and perspectives), and is starting to attract interest from the machine learning community, through quantum algorithms for training models \cite{kerenidis2019q,li2019sublinear,Kerenidis2020QuantumDeep,bausch2020recurrent,yamasaki2020learning,wang2021quantum,you2021exponentially,ostaszewski2021reinforcement,kubler2021inductive}, and also by inspiring improved classical methods \cite{stoudenmire2016supervised,tang2019quantum,panahi2019word2ket,panahi2021shapeshifter}. 

In order to achieve quantum speedup, the choice of the variant of the machine learning problem often matters. For example, the pioneering quantum machine learning method -- quantum Least-Squares Support Vector Machine (LS-SVM)  \cite{RML14} -- focused on quadratic loss and quadratic regularizer, in order to be able to utilize solvers for quantum linear systems of equations, such as HHL \cite{harrow2009quantum}. This and other recent solvers based on quantum manipulation of eigenvalues \cite{somma2016quantum,subramanian2018implementing} can lead to exponential speedup compared to classical methods, as long as the eigenvalues of the linear kernel matrix are of similar magnitude. However, linear systems arise only from unconstrained, or equality-constrained quadratic problems, such as LS-SVM, but not from the standard, hinge-loss SVM. 
A more recent   quantum supervised learning method  \cite{li2019sublinear} that can achieve sublinear training time involves efficient quantum 
primal-dual approach for solving minimax problems, and as a consequence focuses on minimizing the maximum -- not the average -- loss over the training set. For the hinge loss used in the original SVM, a quantum variant of SVM-perf algorithm that  transforms samples from input space to kernel space and then uses quantum methods to approximate inner products in the kernel space, has  been proposed recently  \cite{allcock2020quantum}, achieving complexity linear in the number of samples. 

Here, we focus on the possibility of achieving quantum speedup for the  Sparse SVM \cite{bennett1999combining,kecman2000support, bi2003dimensionality,zhu20041}, a linear classifier that combines the standard SVM hinge loss with the sparsity-promoting $L_1$ regularizer\footnote{See Section \ref{sec:back} for background on binary classification and regularized models.} instead of the standard $L_2$ regularizer. Sparse SVM is useful in cases where the number of features approaches or exceeds the number of training samples, and where interpretability of the classifier matters, not just its predictive abilities. For example, in biomedicine it is important to know if, and how much, each feature contributes to the prediction, and often a small number of features is enough to tell the classes apart. Thus, linear models $h(x;\beta)=\beta^T x $ with weights $\beta, x \in \Rset^p$, which learn a single multiplicative weight $\beta_j$ for each feature $x^j$ of a $p$-dimensional sample $x$, are often preferred over non-linear approaches, such as kernel methods, ensembles, or neural networks. In these settings, we often expect that highly-accurate predictions can be made using just a few discriminative features; we expect that a well-performing model should be sparse, that is, there is a vector $\beta$ composed mostly of zeros that achieves near-optimal accuracy. 
The remaining features either carry no information about the separation of classes, or the information is redundant.

\subsection{Our Contribution}
\begin{figure*}[b]
	\centering
	\includegraphics[width=0.85\textwidth]{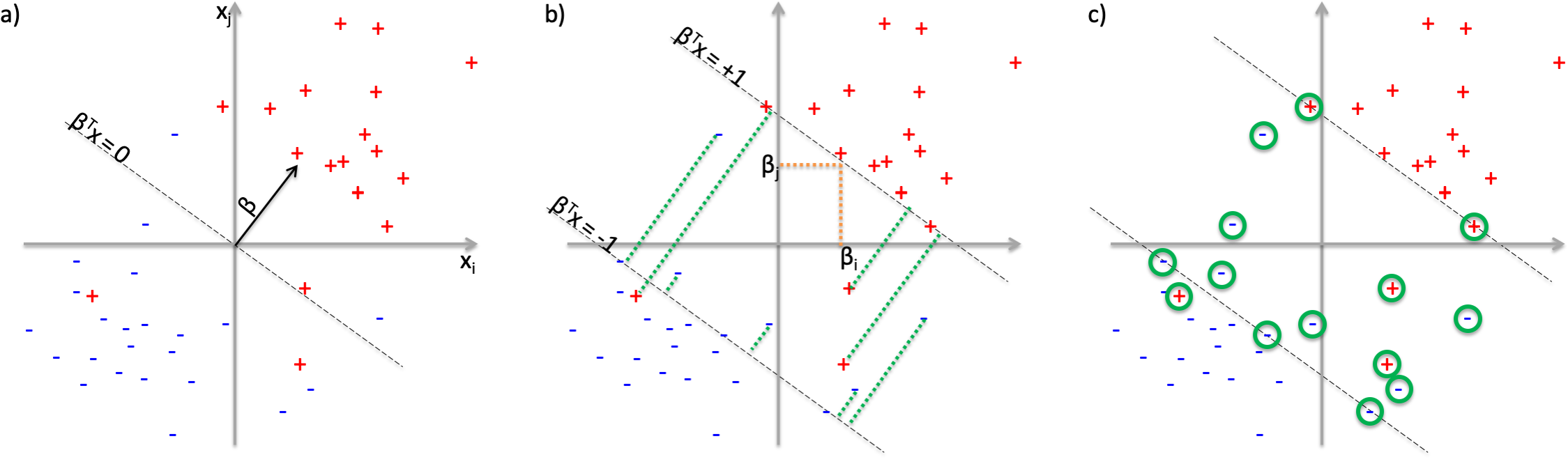}
	\caption{\label{fig:rr} 
		Graphical illustration of the factors contributing to the computational complexity of Quantum Sparse SVM for a classification problem with linear solution vector $\beta$ ({\bf a}). Complexity is proportional to the product $Rr$, where $R$ is the sum of sample loss values ({\bf b}: green lines) and feature weight magnitudes ({\bf b}: orange lines), and $r$ is the sum of weights of support vectors ({\bf c}: green circles).}
\end{figure*}

We propose Quantum Sparse SVM (QsSVM), an approach that minimizes the training-set objective function of the Sparse SVM model \cite{bennett1999combining,kecman2000support, bi2003dimensionality,zhu20041} using a quantum algorithm for solving linear programming (LP) problems 
\cite{van2019quantum} instead of a classical solver. Our aim is to analyze whether using a quantum solver provides benefits in terms of computational complexity of model training, that is, if sublinear training time in terms of the number of training samples, $m$, and the number of features, $p$, be achieved. This problem is challenging, since quantum LP solver \cite{van2019quantum} and similar quantum SDP/LP solvers \cite{brandao2017quantum,van2017quantum,brandao2017large,van2018improvements} express the complexity not only in terms of $m$ and $p$, but also in terms of characteristics of the primal and dual solution to the LP/SDP instance being solved (see Figure \ref{fig:rr}). So far, realistic application scenarios with characteristics that provably lead to quantum speedup for these solvers have been scarce. 

We show that for arbitrary binary classification problems no quantum speedup is achieved using quantum LP solvers. More broadly, we prove that the lower bound for solving Sparse SVM using any quantum algorithm with black-box access to training data is $\Omega(m)$. However, we show there are realistic cases in which a sparse linear model will have high accuracy, and sublinear time in $m$ and $p$ can be guaranteed when using the quantum solver, while proving that any classical algorithm requires at least linear time.

\section{Background on Binary Classification and Sparse Linear Models}
\label{sec:back}

Binary classification involves vector-scalar pairs $(x,y) \in \mathcal{X} \times \mathcal{Y} $, where $\mathcal{Y}=\left\{-1,1\right\}$  and $\mathcal{X} \subset \Rset^p$ is a compact subset of a $p$-dimensional feature space. Each pair describes an object of study, for example a brain scan or a tissue sample of a medical patient. Individual components $x^{j}$ of a vector $x$ are called features. Each feature describes some numerical property of the object represented by $x$, for example signal intensity in a single voxel of a brain scan, or expression level of a single gene. The value of $y$ tells us whether the object belongs to the positive or the negative class. In many scenarios, the feature vectors are easy to obtain, but the class variable is not. For example, we can measure the methylation status of each CpG base-pair in the patient's genome relatively easily, but deciding if the patient's prognosis is positive or negative is challenging. 

In statistical learning \cite{friedman2001elements}, we assume that samples $(x,y)$ come from a fixed but unknown distribution $D$ over $\mathcal{X} \times \mathcal{Y}$. 
For a given feature vector $x$, the probabilities of either class are given by conditional distribution $D_{y|x}$ over $\mathcal{Y}$, and for a given class $y$, the probability density of feature vectors in that class is given by conditional distribution $D_{x|y}$ over $\mathcal{X}$. 
While the underlying distributions $D$, $D_{y|x}$, and $D_{x|y}$ are unknown, we have access to a training set $Z$ consisting of $m$ samples $z_i=(y_i, x_i)$ drawn independently from $D$.
In the binary classification problem the goal is to use the training set to learn how to predict classes $y$ for feature vectors $x$ that are not in the training set. 

The training set can be used to construct a predictive model, in a form of a hypothesis function $h: \mathcal{X} \rightarrow \Rset$, where the sign of $h(x)$ indicates the predicted class for input feature vector $x$. For a given sample $(x,y)$, the prediction is considered correct if the signs of the predicted and the true class agree, that is, if $y h(x)>0$. The predictive model should make as few errors as possible over samples $z=(x,y)$ sampled from distribution $D$, that is, it should minimize $\int_{\mathcal{X} \times \mathcal{Y}} I[yh(x) \leq 0] D(z) \diff z$, where $I$ is an indicator function over Boolean domain returning 1 for true and 0 for false.

A simple but often effective class of hypotheses is the class of linear functions $h(x;\beta,b)=\beta^T x + b=\sum_{j=1}^p \beta_j x^j + b$. A linear predictive model is parameterized by a vector of feature weights $\beta \in \Rset^p$ and a bias term $b\in \Rset$. To simplify the notation, we often add one more dimension to $\mathcal{X}$ with all samples having a value of one. The predictive model is then simply $h(x;\beta)=\beta^T x$, $\beta \in \Rset^{p+1}$, with $\beta_{p+1}$ playing the role of bias. 

Training of a linear model involves finding a suitable parameter vector $\beta$. 
For a single sample $(x,y)$, the suitability of a model $h$ with specific $\beta$ will be captured by a loss function $\ell(y, h(x;\beta))$, which returns a nonnegative real number that we interpret as a measure of our dissatisfaction with the prediction $h(x;\beta)$. 
The natural {\em 0/1 loss}, defined as $\ell(y, h(x;\beta))= I[y \beta^T x \leq  0]$, that is, null for a correct prediction and one otherwise, is not a continuous function of the parameter vector $\beta$, and is flat almost everywhere, leading to problems with finding $\beta$ that minimizes the loss. Instead of the 0/1 loss, a convex function that upper-bounds it is often used in training classification models. For example, the least-square loss $\ell(y,h)=(y-h)^2$ is used in Fisher's Linear Discriminant and in Least-Squares Support Vector Machine (LS-SVM) classifier \cite{suykens1999least}.  However, the least-squares loss, while often used in regression problems, leads to high-magnitude loss if $h(x)$ has large magnitude, even if $h(x)$ and $y$ agree on the sign, that is, the prediction is correct. Most classification loss functions involve a nonincreasing $\Rset \rightarrow \Rset_+$ function of the product $yh$. If the sign of prediction $h(x)$ and the target class $y$ agree, then loss should not increase if $|h|$ increases. One prominent example of a convex, monotonic loss is the {\em hinge loss}, defined as $\ell(y,h)=[1-yh]_{+}=\max(0,1-yh)$, which is  used in the original variant of the Support Vector Machine (SVM) classifier \cite{cortes1995support}. Hinge loss leads to hinge risk $\hat{R}_{SVM}(\beta)=\frac{1}{m}\sum_{i=1}^{m} \max(0,1-y_i \beta^T x_i )$. 

Once the loss function is chosen, the goal of training a model is to find the parameter vector $\beta$ that minimizes the expected loss $\mathbb{E}_{z \sim D} \ell(y, h(x;\beta))$, referred to as {\em risk} of the model, $R(\beta)=\int_{\mathcal{X} \times \mathcal{Y}} \ell(y, h(x;\beta)) D(z) \diff z$. Since $D$ is unknown, a surrogate goal is to search for $\beta$ that leads to low loss on samples from the training set. For example, the {\em empirical risk minimization} strategy involves finding parameters $\beta$ that minimize {\em empirical risk}, that is, the average loss on the training set, $\hat{R}(\beta)=\frac{1}{m}\sum_{i=1}^{m} \ell(y_i, h(x_i;\beta))$. 

The model $\beta$ that minimizes the empirical risk may have high generalization risk $R(\beta)$, that is, may fare poorly on samples outside of the training set, especially if the features are not statistically independent, or if the number of  features $p$ is of the same order or higher than the number of training  samples, $m$. Often, the generalization error can be reduced if a penalty on the complexity of the model is introduced into the optimization problem. Typically, this penalty term, known as regularization term, is based on $\lVert \beta \rVert$, a norm  of the vector of model parameters, leading to {\em regularized empirical risk minimization} strategy, which finds parameters that minimize $\hat{L}(\beta) = \hat{R}(\beta) + \lambda f(\lVert \beta \rVert)$. For example, most Support Vector Machines \cite{cortes1995support} use squared $L_2$ norm of $\beta$, $\lVert \beta \rVert_2^2$, as the regularizer. 

The $L_2$ regularizer used in LS-SVM penalizes large-magnitude feature weights, but is unlikely to set any feature weights to null. In many real-world scenarios involving classification problems with large number of features we expect that highly-accurate predictions can be made using just a few discriminative features. The remaining features either carry no information about the separation of classes, or the information is redundant. For example, classification problems involving gene expression measured using microarrays or RNA-seq may have tens of thousands of features, and brain scans can have million of voxels, but only a small number may be enough to separate subjects with one subtype of a disease from another subtype, an information that is useful in choosing treatment. In these scenarios, we expect that a well-performing model should be sparse; that there is a vector $\beta$ composed mostly of zeros that achieves near-optimal risk $R(\beta)$. 
The key problem is to decide which feature weights should be non-zero. 

To find sparse solutions to classification problems, a regularization term in the form of $L_1$ norm of $\beta$ is often included in the objective function. $L_1$ regularization is especially useful when working with a training set with large number of features compared to the number of training samples, which is referred to as the $p>m$ case. One method that uses $L_1$ regularization is Sparse SVM (sSVM) \cite{bennett1999combining,kecman2000support,zhu20041}, a linear classifier based on regularized empirical risk minimization involving hinge loss and $L_1$ regularizer, $\hat{L}_{sSVM}(\beta) = \hat{R}_{SVM}(\beta) + \lambda \lVert \beta \rVert_1$, where $\lambda > 0$ is a hyperparameter specifying the strength of regularization. Training of a Sparse SVM model can be transformed into an optimization problem with linear objective function and linear inequality constraints. 

\section{Training Sparse SVM using Quantum LP Solvers}

The training of Sparse SVM model using a training set $\left\{(x_i,y_i)\right\}$ with $p$ features and $m$ samples and a regularization constant $\lambda > 0$ involves solving a minimization problem
\begin{align}
\label{eq:SparseSVM}
\argmin_{\beta \in \Rset^p} \frac{1}{m}\sum_{i=1}^{m} \max(0,1-y_i \beta^T x_i ) + \lambda \sum_{j=1}^p | \beta_j |.
\end{align}
Using standard techniques, for $\lambda>0$, this non-linear unconstrained optimization problem can be transformed to an equivalent primal constrained linear program with $n=m+2p$ nonnegative variables and $m$ linear inequality constraints 
\begin{align}
\label{eq:SparseSVMConstr}
\min_{\xi,\beta^+,\beta^-} \;\;\;\; & \frac{1}{m}\sum\limits_{i=1}^{m} \xi_i + \lambda  \sum\limits_{j=1}^{p} \beta_j^+ + \lambda  \sum\limits_{j=1}^{p} \beta_j^-  \\ 
\text{s.t.} \;\;&  \sum\limits_{j=1}^{p} y_i x_i^j \beta_j^+ - \sum\limits_{j=1}^{p} y_i x_i^j \beta_j^-  \geq 1 - \xi_i, \;\;\;\;\; i\in[m]\nonumber \\
\;\; & \xi_i, \beta_j^+, \beta_j^- \geq 0,  \nonumber
\end{align}
where $[m]=\left\{1,...,m\right\}$. We can read out the solution as $\beta_j = \beta_j^+ - \beta_j^-$. We also have $|\beta_j| = \beta_j^+ + \beta_j^-$. The value of the hinge loss of $i$-th training sample is equal to  $\xi_i$. The dual problem is 
\begin{align}
\max_{\alpha}  \sum_{i=1}^m \alpha_i \;\;\;\; 
\textrm{s.t.}   \;\; &-\lambda \leq \sum_i \alpha_i y_i x_i^j \leq \lambda, \;\;\;\;\; j \in [p] \nonumber \\
&\;\;\;\;0 \leq \alpha_i \leq 1/m \;\;\;\;\;  i \in [m]. \label{eq:dualLP}
\end{align}

A classification problem with  $m$ samples in $p$-dimensional feature space leads to a primal LP with $n=m+2p$ nonnegative variables and $m$ linear inequality constraints. 


\subsection{Linear Programs and Matrix Games}
To solve the linear program associated with Sparse SVM, we employ a recently proposed quantum LP solver \cite{van2019quantum} based on zero-sum games. The solver is based on a series of zero-sum matrix games. For a given LP, each game in the series is constructed to indicate if the optimal value of the LP lies in a certain numerical range. 

Each zero-sum matrix game involves two players, Alice with $n$ possible strategies and Bob with $m$ possible strategies. If Alice chooses strategy $i \in [n]$ and Bob chooses $j \in [m]$, the pay-off to Alice is $M_{ij}$ and to Bob is $-M_{ij}$, where $M$ is the matrix defining the game. Given a randomized strategy for Alice, sampled from $u \in \Rset_+^n$, with $\norm{u}=1$, and a randomized strategy for Bob, sampled from $v \in \Rset_+^m$, $\norm{v}=1$, the expected payoff to Alice is $u^TMv$. Bob can assume optimal strategy distribution $u$ on Alice's part and chose his strategy distribution $v$ according to $\min_v \max_u u^TMv$ to minimize Alice's and maximize his own expected pay-off. 

For a given linear program $\mathrm{LP}(c,A,b)=\min c^T x \; \st Ax\leq b, \;x\geq 0$ and a real value $\alpha$, finding the expected pay-off for the optimal strategy in a game defined by a matrix $M$ that contains $\alpha$ as an element and $A$, $b$, $c$ as sub-blocks can be used to determine if the optimal value of the LP is lower than $\alpha$. If it exists, a feasible LP solution with optimal value below $\alpha$ can also be obtained from the optimal strategy vectors $u$ and $v$. Iteratively, this process can solve the LP up to a pre-determined error. 

A classical algorithm \cite{Grigoriadis1994ASR} can solve the game up to $\varepsilon$ additive error in $\BigPolyOh{(n+m)/\varepsilon^2}$ time, where $\tilde{O}$ notation hides logarithmic factors. The quantum version replaces Gibbs sampling step in the classical algorithm with its quantum counterpart, achieving quadratic speedup in terms of $n, m$, and a $1/\varepsilon^3$ dependence on the desired additive error $\varepsilon$ of the solution. A query about the optimal value of an LP with $m$ variables and $n$ constraints being in a certain range reduces to a game with $(n+3) \times (m+2)$ matrix $M$. Further, assuming that the LP primal and dual solution vectors are bounded in $L_1$ norm by $R$ and $r$, respectively, bisection argument shows that only a logarithmic number of such range queries -- matrix games -- are needed, 
each solved to within  $R(r+1)/\varepsilon$ error, in order to obtain the solution to the LP with at most $\varepsilon$ additive error. 
Together, the matrix game-based quantum solver outlined above \cite{van2019quantum} has   $\BigPolyOh{\big(  \sqrt{m}+\sqrt{n}\big) \big( R (r+1)/\varepsilon\big)^3}$ complexity of obtaining the solution to the LP to within $\varepsilon$ additive error, where $\tilde{O}$ notation hides logarithmic factors. The solver achieves a polynomial speedup compared to classical methods, and lower exponent in dependence on $R$, $r$, and $1/\varepsilon$ than in previous quantum methods \cite{brandao2017quantum,van2017quantum,brandao2017large,van2018improvements}.
The key question is whether the dependence on $R$ and $r$, the $L_1$ norms of the primal and dual LP solutions, respectively, allows for achieving sublinear time complexity for  linear programs corresponding to training Sparse SVMs.

\subsection{Data Access Model} 
%
For a linear program $\mathrm{LP}(c,A,b)=\min c^T x \; \st Ax\leq b, \;x\geq 0$, the quantum LP solver requires read-only access to the vectors and matrices $(c,A,b)$ defining the LP, and read/write access to internal data. The solver assumes access to data using a quantum oracle implemented using quantum random access memory (QRAM). For example, the access to element $A_{ij}$ of the LP constraint matrix $A$ is given by an oracle associated with $A$, a unitary linear operator $O_{A}$ capable of performing the mapping 
$O_{A}\ket{i}\Ket{j}\Ket{z} \rightarrow  \ket{i}\ket{j}\ket{z\oplus A_{ij}}$
in superposition. The operator that takes three qubits on input, corresponding to indices $i,j$ and a placeholder $z$, and output produces $i,j$ and the exclusive alternative ($\oplus$) of the binary representation of the matrix element $A_{ij}$ and the placeholder value of the third qubit, $z$. The algorithm assumes the oracle can operate in superposition, that is, given a superposition of indices, returns a superposition of array elements
\begin{align*} 
O_{A}\ket{i}\big(\alpha\Ket{j}+\beta\Ket{j'}\big)\Ket{z} & \rightarrow \alpha \ket{i}\ket{j}\ket{z\oplus A_{ij}} + \beta \ket{i}\ket{j'}\ket{z\oplus A_{ij'}},
\end{align*}
and similarly for the first index. In the LP program for SVM, the matrices and vectors $(c,A,b)$ are derived from input matrix of features $X=[x_i^j]$ and input vector of classes, $y=[y_i]$. Each entry in $A,b,c$ requires only a constant time arithmetic operation on $X$ and $y$, thus, a single oracle call to access $A,b,c$ will require $\BigOh{1}$ calls to oracles for $X$ and $y$.
The computational complexity of the algorithm is measured with respect to number of oracle calls and the number of two-qubit quantum gates required for further processing.

Availability of quantum random access memory is a typical assumption in quantum algorithms, including recent quantum machine learning methods \cite{li2019sublinear,kerenidis2019q,Kerenidis2020QuantumDeep}. Workable, large-scale QRAM  does not exist yet and feasibility of its constructing is still  debated, but algorithmic models  \cite{GLM08,AGCM15,park2019circuit,PhysRevLett.123.250501}  and  experimental demonstrations \cite{parniak2017wavevector,jiang2019experimental,yu2019entanglement,ouyang2020experimental} of quantum memory alone, and as part of quantum networks or quantum learning systems, are emerging. Recent results indicate that loading classical data into quantum RAM can be done in logarithmic
time \cite{zhao2018smooth,larose2020robust}, at the cost of introducing small perturbations into the data.

In a fully-quantum RAM, operations of both reading and writing in quantum superposition should be available. In quantum machine learning systems, typically the input data, as well as certain intermediate data, is processed classically; for example, we assume the feature values for each sample arrive over some classical information channel. Only the read operation is required to operate in quantum superposition, that is, with parallelism implied in linearity of the data access oracle. This type of access is referred to as 
quantum-read, classical-write RAM (QCRAM) \cite{grover2002creating,kerenidis2016quantum}. The standard model of QCRAM utilizes a tree structures over non-zero elements of an $n$-dimensional vector $x$ to allow classical readout and update of a single vector entry $x_i$ in $\BigOh{\log n}$ time. It also allows for accessing selected elements $x_i$ in quantum superposition, sampling integers $i \in [n]$ according to $x_i / \norm{x}$, and creating the quantum state corresponding to $x$.

For a linear program $\mathrm{LP}(c,A,b)=\min c^T x \; \st Ax\leq b, \;x\geq 0$, the quantum LP solver requires read-only access to the vectors and matrices $(c,A,b)$ defining the LP, and read/write access to the strategy distribution vectors $u$, $v$ as it iterates through a sequence of matrix games. The solver assumes access to data via a quantum oracle implemented efficiently using QRAM. Through iterations, the game solution vectors $u,v$ are stored in QCRAM. Solving each game involves an iterative algorithm that result in at most $\BigOh{1/\varepsilon^2 \log mn}$ elements of $u,v$ being non-zero. The QCRAM stores only non-zero elements of $u,v$, thus, accessing all solution elements classically using sequential tree traversal adds $\BigOh{1/\varepsilon^2 \log mn}$ overhead. For sparse models, with few non-zero feature weights in the optimal solution, access can be even faster.

\section{Lower Bound for Complexity of Quantum Sparse SVM}

Assuming efficient oracle access to input, the computational complexity of quantum LP solver utilized in Quantum Sparse SVM shows improved dependence on $n$ and $m$, but polynomial dependence on $R$ and $r$ may erase any gains compared to classical LP solvers. 

For any training set, the minimum of the objective function of the SparseSVM optimization problem (eq. \ref{eq:SparseSVM}) is bounded from above by one, since an objective function value of one can be obtain by setting $\beta = 0$, which leads to unit loss for each training sample, and thus unit average loss. For some training sets, one is the minimum of the objective function -- for example if training samples come in pairs, $(x,+1)$ and $(x,-1)$. In this case, the norm of the primal solution is $R=\sum_i |\xi_i| +\sum_j |\beta_j| = m$, and the norm of the dual solution is $r=\sum_i |\alpha_i| = 1$, since by eq. (\ref{eq:dualLP}) and strong duality it is equal to the value of the primal objective function. 
The quantum solver we use includes $(R (r+1)/\varepsilon)^3$ term in its complexity, and $Rr=\BigOh{m}$ erases any speedups compared to classical solvers. 

A more realistic case in which we see  $R=\BigOh{m}$ is a regular XOR problem, for example involving two features and four training samples, $[+1,+1]$ and $[-1,-1]$ with $y=+1$ and $[+1,-1]$, $[-1,+1]$ with $y=-1$. For any $\beta \in \Rset^2$, if there is a sample with loss $1-\delta$, there is another sample with loss $1+\delta$. Thus, sum of $\xi_i$ variables is one for any $\beta$, and again $\beta=0$ is the minimizer of the regularized empirical risk, leading to $Rr=\BigOh{m}$ and a $\BigOh{m^{3.5}}$ term in the solver worst-case computational complexity. 

The above negative results concern speedup of Sparse SVM utilizing a specific quantum LP solver. It can be shown that sublinear worst-case complexity, in terms of the smaller of $m$ and $p$, is not possible in general. 
To provide lower bound on solving Sparse SVM using any quantum algorithm with black-box access to elements of feature vectors $x$ and class variables $y$, we utilize reduction from majority problem, that is, the problem of finding the majority element in a binary vector $\majV$ of size $n$. In the majority problem, vector $\majV$ has $t$ unity and $s=n-t$ null elements, and the algorithm should return true if $t>s$ or false otherwise. Alternatively, it suffices to return the value of $t$ or $s$. 

Given arbitrary majority problem instance, we show a simple procedure to treat it as the input training set for a Sparse SVM classifier. The procedure introduces no overhead, one call to the training data oracle translates to $\BigTheta{1}$ computation and at most one call to the underlying majority problem oracle. We prove that solving the Sparse SVM classification problem, for values of the regularization constant $\lambda$ leading to non-null feature weights vector $\beta$, allows us to provide the answer to the underlying instance of the majority problem. The lower bound for computational complexity of the majority problem, assuming black-box quantum oracle access to elements $\majV_i$ in superposition, is $\Omega(n)$ \cite{beals2001quantum}, leading to the lower bound on the complexity of training Sparse SVM.

\begin{theorem}
	\label{thm:lowerBound}
	Assuming black-box quantum oracle access to a training set with $m\geq8$ samples and $p$ features,  the lower bound for training Sparse SVM, for regularization constants $\lambda$  that do not lead to all-null models $\beta=0$, has complexity $\BigOmega{m}$, even if the optimization algorithm returns  a suboptimal solution with objective value $ \hat{L} = L^*+ \BigOh{1}$.
	\begin{proof} 
		For given $m\geq 8$\footnote{We use reduction to majority problem of size $n=\floor{m/2}$. For a majority problem to be non-trivial, we need $n\geq 3$, leading to $m\geq 6$ requirement. The requirement of $m\geq 8$ is only needed for the case of approximate solver, returning suboptimal solution.}, $p \geq 1$, set $n=\floor{m/2}$; we have $2n \leq m\leq 2n+1$, and $n\geq 4$. Let $\majV_{n,t}$ be an arbitrary instance of a majority problem of size $n$ with $1 \leq t \leq n- 1$ ones and $n-t$ nulls.
		We  construct an $m \times p$ training set based on  majority problem instance $\majV_{n,t}$  in the following way. 
		
		We start with defining values for the first feature, $j=1$, for all $m$ samples. For $i \in \left\{1,...,n\right\}$, the samples belong to the negative class, that is, $y_i=-1$. For obtaining their feature values $x_i^{j=1}$, we rely on the underlying majority problem. We set $x_i^{j=1}=1$ if $\majV_{n,t}(i)=1$, and set $x_i^{j=1} = -1$ if $\majV_{n,t}(i)=0$. We define samples $i=n+1,...,m$ to have the same, unit value of the first feature, $x_i^{j=1}=1$;  these samples all belong to the positive class, $y_i=1$. For the remaining features $j=2,...,p$, we assign null value for all $m$ samples, that is, for $j\geq 2$, $x_i^j=0$ for any $i\in [1,m]$. 
		
		The $m \times p$ training set is not constructed explicitly. Instead, the oracle access to its elements involves determining, using indices $i,j$ and $\BigTheta{1}$ computation, whether a single call to the underlying majority problem black-box oracle is required (for $j=1$, $i=1,...,n$)  or whether to return a pre-determined value without the need for accessing the underlying oracle. Thus, the number of oracle calls to the training data  is bounded from below by the number of calls to the underlying majority problem oracle.

		For the training problem instance defined above, in the optimal Sparse SVM solution\footnote{We use the notation $[\cdot]_+ = \max(0,\cdot)$.} $\argmin_{\beta \in \Rset^p} \frac{1}{m}\sum_{i=1}^{m} [1-y_i \beta^T x_i ]_+ + \lambda \sum_{j=1}^p | \beta_j |$, features $j=2,...,p$ have null model weight $\beta_j=0$, since the feature values $x_i^j$ for $j\geq 2$ are all zeros and do not contribute anything to the classifier decision irrespective of $\beta_j$ value, while $L_1$ regularization penalizes for non-zero values of feature weights. 
		
		We turn our focus to optimal model weight for the first feature, $\beta_1$. Consider $\beta_1=1$. The $t$ samples corresponding to $\majV_{n,t}(i)=1$ will be misclassified (prediction $1$, true class $-1$), and each will contribute $[1-y_i \beta_1 x_i^1 ]_+/m = 2/m$ loss to the  objective value. The $n-t$ samples corresponding to $\majV_{n,t}(i)=0$, as well all the samples $n+1,...,2n$, will be classified correctly (prediction vs. true class is  $-1$ vs. $-1$ or $1$ vs. $1$), contributing null loss. Regularization term will contribute $\lambda |\beta_1| = \lambda$. In total, we will have $L = 2t/m + \lambda$. 
		
		Any negative $\beta_1$ has higher objective value than the same-magnitude positive $\beta_1$, allowing us to focus on nonnegative values of $\beta_1$. Increasing $\beta_1$ above 1 will increase the loss on the $t$ samples and keep null loss for the $2n-t$ samples, and will increase the regularization term. Having $\beta_1 = 1-\delta$ for $0 < \delta \leq 1$ will reduce loss on the $t$ samples by $\delta/m$ per sample, increase loss on the $2n-t$ samples by the same amount, and decrease the regularizer term by $\lambda \delta$, for a total change of $(2n-t)\delta/m - t\delta/m - \lambda \delta = \delta(2n-2t - \lambda m) /m$ in the objective value. 
		
		For regularization constants $\lambda \geq 2(n-t)/m $, minimal objective value is attained by setting $\beta_1=0$, that is, by an uninformative, all-null $\beta=0$ model. For any $\lambda < 2(n-t)/m$, the optimal model is $\beta_1=1$, $\beta_{\geq 2}=0$, with optimal objective value $L^*  = 2t/m + \lambda$. 
		
		For the non-all-null models, the observed optimal objective value $L^*$ can be used to solve the underlying majority problem by calculating $t$ using 
		\begin{align*}
		t &= m (L^* - \lambda)/2.
		\end{align*}
		Hence, among datasets with $m$ samples and $p$ features are those corresponding to all possible majority problems of size $n=\floor{m/2}$, some of which require number of oracle calls proportional to $n$ to be solved. Thus, quantum Sparse SVM is $\BigOmega{m}$.
		
		If the solver returns a suboptimal solution with objective value $\hat{L}=L^* + \varepsilon$ and we use it to obtain $\hat{t} =m (\hat{L} - \lambda)/2$, we make an error
		\begin{align*}
		\Delta t = \hat{t}-t & =m\varepsilon/2. 
		\end{align*} 
		
		Consider a promise variant of majority, in which the minority class is guaranteed to have at most $k$ elements, which has $\BigOmega{k}$ complexity lower bound. For problems with the $k=n/4$-minority class promise, $n/4$ oracle calls are needed in the worst case; furthermore, making smaller than $n/4$ error $\Delta t$ in estimating $t$ does not change the correctness of the solution.
		
		To have the error $\Delta t < n/4$ that guarantees no mistakes in solving the $n/4$-promise majority problem, we need 
		\begin{align*}
		\varepsilon &< n/2m.
		\end{align*}
		Noting that $2n \leq m\leq 2n+1$ and that $m\geq 8$, the inequality  is guaranteed if 
		\begin{align*}
		\varepsilon \leq 1/5.
		\end{align*}
		Any suboptimal solution with additive error of  $1/5$ or less compared to the optimal objective value allows us to solve $n/4$-promise majority problems, some of which require $\BigOmega{n}$ oracle calls. Thus, allowing $\BigOh{1}$ solver error still leads to $\BigOmega{m}$ complexity lower bound. 
	\end{proof}
\end{theorem}
Note that $L^* \leq 1$, since $L^*=1$ can always be achieved irrespective of the training set by an all-null feature weights vector $\beta$. The error in estimating $L^*$  within which it is still possible to solve the promise majority problems does not decrease with  the input problem size, neither in absolute terms, nor in relative terms when we focus on multiplicative error $\rho=(L^* + \varepsilon ) /L^*$ instead of $\varepsilon$.

\section{Speedup in Quantum Sparse SVM for Special Cases}

Worst-case analysis of the proposed method utilizing quantum LP solver \cite{van2019quantum}, and of quantum Sparse SVM in general, shows that it cannot achieve sublinear computational complexity. This does not preclude sublinear complexity for some families of classification problems for specific quantum SVM solvers. 

In the quantum LP-based approach, sublinear time can only be achieved if the $L_1$ norms of the primal and dual solution vectors are kept in check as the number of samples and features grows. The regularizing term $\norm{\beta}_1$ can be expected to grow slowly in sparse models, and the average loss is normalized by the number of training samples and thus always below one. Yet, sparsity assumption alone is not enough to achieve speedup when class distributions overlap in the features space and the optimal error is not null. While the bound on the dual solution norm is a direct consequence of sparsity of the model, the $\sum_i \xi_i$ term in the primal solution norm in principle grows in proportion to the number of training samples and to the generalization risk, and may be further inflated by the stochastic nature of the training set. We show that under realistic assumptions about the family of classification problems for growing number of features, where new features occasionally bring new discriminative information, a bound on the primal solution norm that leads to sublinear training time can be formulated. 

\subsection{Truncated Subgaussian Classification Problems}

The first defining characteristic of the family of classification problems we explore here is limited overlap between class distributions. Since we are focusing on a linear, sparse SVM classifier, the overlap will be with respect to a linear decision boundary -- we will assume that the tails of class conditional distributions that reach across the decision hyperplane are bounded, in a way formalized in Definition \ref{def:trunc}. The consequences of that assumption\footnote{In this section we motivate Definition \ref{def:trunc} and present the  final complexity result in Theorem \ref{thm:softFin}, and we defer the detailed Lemmas to Section \ref{sec:ComplexityLemmas}} on the expected loss of the model are analyzed in Lemma \ref{lem:ExpVar} in the distribution setting, and then extended in Lemma \ref{lem:riskBound} to the  empirical risk on finite-sample datasets of fixed size $(m,p)$ that are actually seen during training, providing basis for asymptotic analysis building up to Theorem \ref{thm:softFin}. 

In the asymptotic analysis of time complexity, we will assume that we are given a series of classification problems where both the number of samples, $m$, and the number of features, $p$, grows. 
The second defining characteristic of the scenarios leading to sublinear time complexity is the assumption that $p'$, the number of discriminative features used by the sparse model, also grows, but at a slower rate than $m$ or $p$. 
In an illustrative idealized case, these discriminative features have means $+c$ and $-c$ in the positive and the negative class, respectively -- though the situation does not change if the signs of the means are swapped for some of the discriminative features. Thus, $c$ represents how discriminative a feature is. With increasing number $p'$ of discriminative features, each with means differing by at least $2c$, the distance between the means of the two multivariate distributions increases at the rate of at least $2c\sqrt{p'}$. Moving beyond the idealized case of means of each feature at $\pm c$, we will simply require that the multivariate class distribution means diverge at this rate. This defining feature of our scenario is key to asymptotic analysis starting in Lemmas \ref{lem:emrisk} and \ref{lem:softRR} and concluding with  the final complexity result in Theorem \ref{thm:softFin}.

\begin{figure*}[t]
	\centering
	\includegraphics[width=0.95\textwidth]{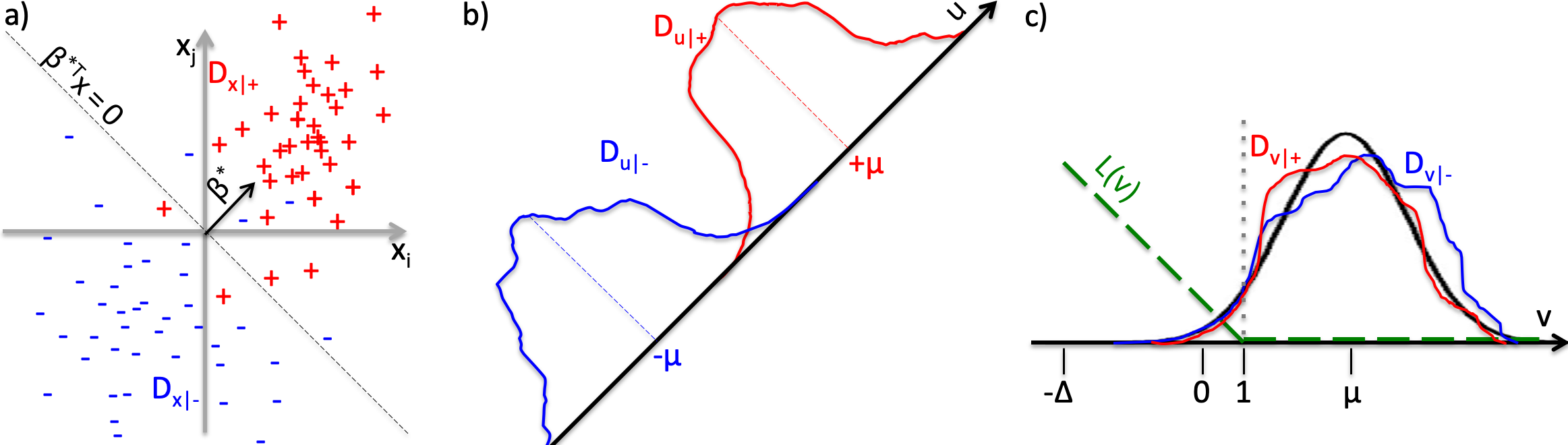}
	\caption{\label{fig:softMargin} 
		Graphical illustration of Definition \ref{def:trunc}: {\bf a)} the initial multivariate two-class problem; {\bf b)} intermediate step involving projection $u={\beta^*}^Tx$ into one dimension; {\bf c)} final univariate distributions after projection $v=yu$.}
\end{figure*}

\begin{mydef}
	\label{def:trunc}
	A {\em $(\Delta,\mu)$-truncated subgaussian classification problem}, for $\mu>1$, $\Delta>0$, is defined by distribution $D$ such that there is an underlying vector $\beta^\sstar \in \Rset^p$ with $\lVert \beta^\sstar \rVert_2 =1$, for which
	\begin{itemize}
		\item the conditional distributions $D_{x|+}$ and $D_{x|-}$ of the samples from the positive and negative class, respectively, give rise to univariate distributions $D_{v|+}$ and $D_{v|-}$ on a line resulting from the projection $v=y{\beta^\sstar}^T x$, 
		\item the tails of $D_{v|+}$ and $D_{v|-}$ are bounded from above, in the region $v \in (-\infty,1]$, by the normal probability density function $\mathcal{N}_{\mu,1}(v)$ with unit standard deviation, centered at $\mu$,
		\item the tails of $D_{v|+}$ and $D_{v|-}$ have zero mass for $v<-\Delta$.
	\end{itemize}
	A $p$-dimensional $(\Delta,\mu)$-truncated subgaussian problem is called sparse if the number of non-zero components in the vector $\beta^\sstar$ is small compared to the number of features, $p$. 
\end{mydef}
A graphical illustration of the definition is shown in Figure \ref{fig:softMargin}.
Sparse $(\Delta,\mu)$-truncated subgaussian problems often arise in dataset coming from natural science. For example, in molecular biology, levels of raw mRNA transcripts of individual genes approximately follow a log-normal distribution within a class of samples \cite{bengtsson2005gene,ntranos2019discriminative}. Raw data is typically processed via a log-like transform prior to analyses, then each class of a multi-gene dataset approximately follows a multivariate normal distribution, leading to sub-Gaussian tail. Biomedical data often has large number of features, and the samples are rarely sparse, that is, vast majority of feature values are non-zero.  For example, classification problems involving gene expression measured using RNA-seq may have tens of thousands of features, only a relatively small number of genes contribute to between-class differences, leading to a sparse problem -- sparse linear models have been successful in analyzing  brain activity \cite{liu2019human}, microbiome \cite{fettweis2019vaginal}, and in other biomedical settings \cite{davies2017hrdetect,ackerman2018route,cheung2018single}. These application scenarios also typically lead to  datasets in which the number of features is larger or at least comparable to the number of samples, $p \gtrsim m$. 

We focus on the  $p \gtrsim m$ scenarios common in differential gene expression or methylation analysis, or classification of 3D brain scans, in which $L_1$ regularization is especially useful. Training the Quantum Sparse SVM for sparse $(\Delta,\mu)$-truncated subgaussian  problems, assuming datasets with $p \gtrsim m$, has complexity that is sublinear in terms of number of samples $m$ and number of features $p$.
\begin{theorem}
	\label{thm:softFin}
	For $p \rightarrow \infty$, consider a family of $p$-dimensional $(\Delta_p,\mu_p)$-truncated subgaussian  problems $D_p$ with underlying vectors $\beta_p^\sstar$. Assume that the vector $\beta_p^\sstar$ is sparse, it only has $p' = 1+8 \log p$ non-zero coefficients. Further, assume that the mean $\mu_p$ diverges with the number of discriminative features $p'$ as $\mu_p > c\sqrt{p'}$ for some $c>1$ that captures how separated class centers are for individual informative features. With only $p' << p$ features being informative, we further assume that $\mu_p$ diverges sublinearly in $p$, $\mu_p < p$. As $p$ grows, we allow scattering of the samples farther into the region dominated by the other class, $\Delta_p \leq 8\log p$.  For growing $p$, assume efficient oracle access to the training data, with the training set sizes $m$ growing proportionally with $p$,  $m/p = \BigOh{1}$. Then, training QsSVM has computational complexity of 
	\begin{align*}
	\BigPolyOh{ \sqrt{m+2p}\; \mathrm{poly}\big(\log p,1/\varepsilon\big)}.
	\end{align*}
	\begin{proof} 
		Training QsSVMs translates to solving an LP problem (eq. \ref{eq:SparseSVMConstr}) with $m$ constraints and $n=2p+m$ variables. The quantum LP solver proposed of van Apeldoorn and Gily{\'e}n \cite{van2019quantum} has complexity $\BigPolyOh{\big(  \sqrt{m}+\sqrt{n}\big) \big( R_p (r_p+1)/\varepsilon\big)^3}$. When $m/p = \BigOh{1}$, by virtue Lemma \ref{lem:softRR} (see Section \ref{sec:ComplexityLemmas}), neither $R_p$ nor $r_p$ grow with $m$, and both grow with $p$ as $\BigOh{\log p}$, yielding the complexity result.
	\end{proof}
\end{theorem}

\subsection{Speedup Compared to Classical Sparse SVM  for Truncated Subgaussian Problems}

We analyze here how the sublinear complexity of Quantum Sparse SVM for truncated subgaussian problems compares to classical algorithms. Specifically, we show that classical algorithms for this class of problems cannot be faster than linear, through reduction from the search problem. As a result, Quantum Sparse SVM holds polynomial complexity advantage over any classical algorithm for Sparse SVM on this class of problems. 

Consider the following truncated subgaussian problem with $m$ samples and $p$ features. All samples from the positive class are vectors with $p$ ones. On the other hand, all samples from the negative class have negative one on the same subset of $p'=\BigOh{\log p}$ features, and have the value of positive one for all other features. This setup meets the criteria for a sparse truncated subgaussian classification problem according to Definition \ref{def:trunc}. Finding the optimal classification model involves finding at least one of the $p'$ features where a positive sample differs in value from a negative sample, out of $p$ possibilities. Successfully training the classifier for this particular classification problem translates to successfully solving the search problem. For small values of the $p'/p$ ratio considered here, the search problem has classical complexity of $\BigOh{p}$, providing lower bound on classical algorithms for training Sparse SVMs for truncated subgaussian problems. In the quantum setting, the problem can be solved by Groveer's search algorithm, which attains the quantum lower bound of  $\BigOh{\sqrt{p}}$. The quantum lower bound shows that the proposed method involving a quantum LP solver is optimal, up to the slower growing polylogarithmic terms. 

We also note that since all the sample feature vectors are composed of elements with the same, unit magnitude, and thus all rows and all columns have the same norm,  it is not possible to use norm-based sampling methods to construct a dequantized version of Quantum Sparse SVM with sublinear complexity for the truncated subgaussian problems.

\subsection{Complexity Analysis of Quantum Sparse SVM for Truncated Subgaussian Problems}
\label{sec:ComplexityLemmas}

The path from Definition \ref{def:trunc} to Theorem \ref{thm:softFin} leads through a series of technical Lemmas, from generalization risk to the characteristics of the empirical solution.

For the hinge loss, the generalization risk $R({\beta^\sstar})$ associated with model $h(x)={\beta^\sstar}^T x$ on the $(\Delta,\mu)$-truncated subgaussian problem $D$ is bounded through the following lemma.
\begin{lemma}
	\label{lem:ExpVar}
	Let $D$ be a $(\Delta,\mu)$-truncated subgaussian classification problem with underlying vector  $\beta^\sstar$ leading to univariate distributions $D_{v|+}$ and $D_{v|-}$ as described above. Let $L=\max(0,1-v)$ be a univariate random variable capturing hinge loss of the model $h(x)={\beta^\sstar}^T x$ for samples from $D$. Then, the expectation and standard deviation of $L$ are bounded from above by 
	\begin{align}
	&R({\beta^\sstar}) = \mathbb{E}[L] \leq  \frac{1}{\sqrt{2\pi}}\me^{-\frac{(1-\mu)^2}{2}} = \mathcal{N}_{\mu}(1), \label{eq:EL}\\
	&\mathbb{V}\mathrm{ar}[L] \leq  \left[(1-\mu)^2+1\right]\left[1+\erf\left(\frac{1-\mu}{\sqrt{2}}\right)\right] \label{eq:VarL}.
	\end{align}
	Also, values of $L$ are in the range $[0,\Delta+1]$. 
	\begin{proof}
		The proof relies on properties of integrals of $x^k \mathcal{N}_{0}(x)$. 
		Let 
		\begin{align*}
		G(x)&=\mathcal{N}_{0}(x)= \frac{1}{\sqrt{2\pi}} \me^{-\frac{x^2}{2}}, \;\; G_k(x)= \int_{-\infty}^x t^k  G(t) \diff t.
		\end{align*}
		Then, we have
		Let $p_+$ and $p_-$ by the probabilities, under $D$, of the positive and the negative class, respectively. We have 
		\begin{align*}
		\mathbb{E}[L] & =  \int_{-\infty}^{\infty} \max(0,1-v)  [ p_{+} D_{v|+}(v) + p_{-} D_{v|-}(v) ] \diff v \nonumber \\
		&\leq   (1-\mu) G_0(1-\mu) - G_1(1-\mu)  \leq \frac{1}{\sqrt{2\pi}}\me^{-\frac{(1-\mu)^2}{2}}  \nonumber. \\
		\mathbb{E}[L^2] & \leq   [(1-\mu)^2+1] G_0(1-\mu)  +(1-\mu) G(1-\mu)\\
		& \leq    \left[ (1-\mu)^2+1 \right]  \left[ 1+\erf\left(  \frac{1-\mu}{\sqrt{2}}  \right)  \right] \nonumber.
		\end{align*}
		The range of $L$ follows immediately from null mass of $D_{v|y}$ for $v\leq -\Delta$, and from null loss for any $v\geq 1$.  
	\end{proof}
\end{lemma}

The result above gives the bound on the expected value of the hinge loss for the model $h(x)={\beta^\sstar}^T x$ on the distribution $D$, that is, it bounds from above the generalization risk of that model, $R({\beta^\sstar})=\mathbb{E}[L]$. However, it does not give an upper bound on the empirical risk for the model $h(x)={\beta^\sstar}^T x$ on a specific training set with $m$ samples and $p$ features, sampled from $D$. This bound is given be the following lemma. 
\begin{lemma}
	\label{lem:riskBound}
	Let $D$ be a $(\Delta,\mu)$-truncated subgaussian problem based on $\beta^\sstar$. Let $\hat{R}({\beta^\sstar})$ be the empirical risk associated with model $\beta^\sstar$ over a $m$-sample training set sampled {\em i.i.d.} from $D$. Then, with probability at least $1-\delta$
	\begin{align}
	\hat{R}({\beta^\sstar}) &\leq \frac{1}{\sqrt{2\pi}}\me^{-\frac{(1-\mu)^2}{2}} + 4 \frac{ (\Delta+1) \log(2/\delta)}{m} \label{eq:RhatBound}\\
	&+  4 \frac{  \sqrt{\log(2/\delta)}}  {\sqrt{m}}  \left[ (1-\mu)^2+1\right]\left[1+\erf\left(\frac{1-\mu}{\sqrt{2}}\right) \right]    \nonumber 
	\end{align}
	\mute{	\begin{align}
		\hat{R}({\beta^\sstar}) &\leq \frac{1}{\sqrt{2\pi}}\me^{-\frac{(1-\mu)^2}{2}} \label{eq:RhatBound}\\
		&+  4 \frac{  \sqrt{\log(2/\delta)}}  {\sqrt{m}}  \left[ (1-\mu)^2+1\right]\left[1+\erf\left(\frac{1-\mu}{\sqrt{2}}\right) \right]  \nonumber \\
		&+ 4 \frac{ (\Delta+1) \log(2/\delta)}{m}  \nonumber 
		\end{align}}
	\begin{proof}
		Consider $m$ values $l_1, ..., l_m$ drawn from a univariate random variable $L$ taking values in range in $[a,b]=[0,\Delta +1]$, and with finite variance $s=\mathbb{V}\mathrm{ar}[L]$ and finite mean $R=\mathbb{E}[L]$. Let $\hat{R}=\frac{1}{m}\sum_{i=1}^{m} l_i$ be the empirical mean. Bernstein's inequality states that 
		\begin{align*}
		\mathbb{P}( | \hat{R} - R | \geq t ) \leq  2 \exp\left( -\frac{mt^2}{2(s^2 +(b-a) t)} \right).
		\end{align*}
		That is, with probability at least $1-\delta$, 
		\begin{align*}
		\hat{R} \leq R +   4s\sqrt{\frac{\log(2/\delta)}{m}} + \frac{4(b-a) \log(2/\delta)}{m}.
		\end{align*}
		We thus have
		\begin{align*} \hat{R}({\beta^\sstar}) \leq \mathbb{E}[L] +  4 \mathbb{V}\mathrm{ar}[L] \sqrt{\frac{\log(2/\delta)}{m}} + \frac{4 (\Delta+1) \log(2/\delta)}{m}
		\end{align*}
		The bound follows from plugging in the bounds on expected value (eq. \ref{eq:EL}) and variance (eq. \ref{eq:VarL}) of the loss. 
	\end{proof}
\end{lemma}

We are now ready to analyze the behavior of empirical risk of models $\beta_p^\sstar$ on problems $D_p$ as the number of all features $p$ and the number of discriminative features $p'$ grow.  
\begin{lemma}
	\label{lem:emrisk}
	For $p \rightarrow \infty$, consider a family of $p$-dimensional $(\Delta_p,\mu_p)$-truncated subgaussian  problems $D_p$ with underlying vectors $\beta_p^\sstar$. Assume that the vector $\beta_p^\sstar$ is sparse, it only has $p' = 1+8 \log p$ non-zero coefficients. Further, assume that the mean $\mu_p$ diverges with the number of discriminative features $p'$ as $\mu_p > c\sqrt{p'}$ for some $c>1$ that captures how separated class centers are for individual informative features. With only $p' << p$ features being informative, we further assume that $\mu_p$ diverges sublinearly in $p$, $\mu_p < p$. As $p$ grows, we allow scattering of the samples farther into the region dominated by the other class -- specifically, we allow $\Delta_p \leq 8\log p$.  Then, with probability at least $1-\delta$, we have 
	\begin{align}
	\label{eq:RhatBound2}
	\hat{R}({\beta_p^\sstar}) &\leq  \frac{1}{\sqrt{2\pi} p }+ 4 \frac{ (8\log p+1) \log(2/\delta)}{m} + \BigOh{\frac{1}{\sqrt{m}p^2}}.
	\end{align}
	\begin{proof}
		
		We start with eq. (\ref{eq:RhatBound}) and bound the first and third terms.
	
	For the first term in eq. (\ref{eq:RhatBound}), we will use the following limit $\lim_{x\rightarrow \infty} c\sqrt{1+k x} / [1 + \sqrt{kx}] = c$. Thus, under the assumption that $\mu_p$ grows at least as $c\sqrt{p'}=c\sqrt{1+8\log p}$, for $c>1$, for sufficiently large $p$,  we have
		$\mu_p \geq c\sqrt{1+8 \log p} \geq 1 + \sqrt{8\log p}$, that is, $1-\mu_p \leq -\sqrt{8 \log p}$. Both terms are negative, thus we also have $(1-\mu_p)^2 \geq 8 \log p$, and $-(1-\mu_p)^2/2 \leq -4 \log p$. Since $\me^{x}$ is an increasing function of $x$,
		we thus have  the following upper bound
		\begin{align*}
		\frac{1}{\sqrt{2\pi}}\me^{-\frac{(1-\mu_p)^2}{2}}  \leq \frac{1}{\sqrt{2\pi}}\me^{-4 \log p} = \frac{1}{\sqrt{2\pi}p^{4}} \leq \frac{1}{\sqrt{2\pi}p} \nonumber.
		\end{align*}. 
		
		For the third term in eq. (\ref{eq:RhatBound}), 
		\begin{align*}
		 4 \frac{  \sqrt{\log(2/\delta)}}  {\sqrt{m}}  \left[ (1-\mu)^2+1\right]\left[1+\erf\left(\frac{1-\mu}{\sqrt{2}}\right) \right],
		\end{align*}
		we first observe that, since $\mu_p < p$, the $(1-\mu)^2+1$ in the product is bounded by $p^2$. We will then show that the second term goes to null at least as fast as $1/p^4$, and thus the third term in eq. (\ref{eq:RhatBound}) is $\BigOh{\frac{1}{\sqrt{m}p^2}}$.
		Since $\frac{1-\mu_p}{\sqrt{2}} \leq -\sqrt{4 \log p}$ and $\erf(x)$ is an increasing function of $x$, we have  $1+\erf\left(\frac{1-\mu_p}{\sqrt{2}}\right) \leq 1+\erf(-\sqrt{4 \log p}) $.
		To show that this term goes to null at least as fast as $1/p^4$, we will use L'H\^{o}pital's rule. We first observe that 
		\begin{align*}
		\frac{\diff \left[ 1+\erf(-\sqrt{4
		 \log p})\right] }{\diff p} = -\frac{2 }{p^{5}\sqrt{\pi\log p}}.
		\end{align*}
		From the L'H\^{o}pital's rule, we have
		\begin{align*}
		\lim_{p \to \infty} \frac{1+\erf(-\sqrt{4 \log p})}{p^{-3}}= \lim_{p\to \infty }   -\frac{2 }{p^{5}\sqrt{\pi\log p}}\frac{-4p^5} =\lim_{p\to \infty \frac{8}{\sqrt{\pi\log p}}} =0 \nonumber.
		\end{align*}
		Thus, $\left[ (1-\mu_p)^2+1\right]\left[1+\erf\left(\frac{1-\mu_p}{\sqrt{2}}\right) \right] = \BigOh{ p^2 / p^4 } = \BigOh{ 1/p^2}$. 
	\end{proof}
\end{lemma}

Sparse SVM involves regularized empirical risk, that is, minimization of a weighted sum of the empirical risk and the $L_1$ norm of the model $\beta$. 
Under the scenario of slowly increasing number of discriminative features, the Sparse SVM regularized empirical risk minimization is characterized by the following lemma. 
\begin{lemma}
	\label{lem:softRR}
	For $p \rightarrow \infty$, consider a family of $p$-dimensional classification problems $D_p$ as described in Lemma \ref{lem:emrisk}. For each $D_p$, consider the SparseSVM regularized empirical minimization problem (eq. \ref{eq:SparseSVM}) 
	\begin{align*} 
	\argmin_{\beta} &\spc \frac{1}{m}\sum\limits_{i=1}^{m} \max(0,1-y_i \beta^T x_i) + \lambda \lVert \beta \rVert_1,
	\end{align*}
	involving $m$-sample training set sampled from $D_p$.
	Then, for each $p$, with probability $1-\delta$, there exist an empirical minimizer $\hat{\beta_p}$ of the problem above that can be found using a linear program (eq. \ref{eq:SparseSVMConstr}), with $L_1$ norms of the primal and dual solutions, $R_p$ and $r_p$, respectively, bounded from above by 
	\begin{align*}
	R_p \leq \frac{1}{\sqrt{2\pi}} \frac{m}{p} +  4 (1 + 8 \log p) \log(2/\delta) ] + \lambda \sqrt{1+8 \log p}, \\ 
	r_p \leq \frac{1}{\sqrt{2\pi}} \frac{1}{p} +  \frac{ 4(1 + 8 \log p )\log(2/\delta)}{m} + \lambda \sqrt{1+8 \log p} .
	\end{align*}
	\begin{proof}
		For any unit $L_2$-norm vector $\beta$ with $p'$ non-zero entries, the highest $L_1$ norm is achieved if all the $p'$ coordinates are equal to $1/\sqrt{p'}$. Thus, $\lVert \beta_p^\sstar \rVert_1 \leq \sqrt{p'}$ for $\beta_p^\sstar$ with unit $L_2$-norm and at most $p'$ non-zero coefficients. 
		On the training set, the minimizer $\hat{\beta_p}$ has lowest objective function of all possible $\beta$, including $\beta_p^\sstar$. Together with Lemma \ref{lem:emrisk}, where we omitted the lower order $\BigOh{1/\sqrt{m}p^2}$ term that can be folded into higher-order terms, we have
		\begin{align*}
		&\hat{R}(\hat{\beta_p}) + \lambda \lVert \hat{\beta_p} \rVert_1 \leq 
		\hat{R}({\beta_p^\sstar}) + \lambda \lVert \beta_p^\sstar \rVert_1 \nonumber \\
		&\leq  \frac{1}{\sqrt{2\pi}} \frac{1}{p} + 4 \frac{ (1 + 8 \log p) \log(2/\delta)}{m} + \lambda \sqrt{1+8 \log p} \nonumber.
		\end{align*}
		From strong duality, we have $r_p = \hat{R}(\hat{\beta_p}) + \lambda \lVert \hat{\beta_p} \rVert_1$. The norm $R_p$ of the primal solution does not involve averaging the losses $\max(0,1-y_i \beta^T x_i)$. Instead, the losses are added up, that is, $R_p$ includes the term $m\hat{R}(\hat{\beta_p})$ instead of the empirical risk.
	\end{proof}
\end{lemma}
\noindent Lemma \ref{lem:softRR} directly leads to Theorem \ref{thm:softFin}.

\section{Related Work}

Considerable effort has been devoted into designing fast classical algorithms for training SVMs, although the focus is on the traditional SVM that involves the $L_2$ regularizer, which results in a strongly convex objective function even for hinge loss, as opposed to Sparse SVM, where the objective function is piece-wise linear, and thus convex but not strongly convex.  
The decomposition-based methods such as SMO \cite{platt1998sequential} are able to efficiently manage problems with large number of features $p$, but their computational complexities are super-linear in $m$.  Other training strategies \cite{suykens1999least,fung2005multicategory,keerthi2005modified} are linear in $m$ but scale quadratically in $p$ in the worst case. 
In more recent work, the number of features $p$ is replaced by the average sparsity of samples, $s$, that is, the average number of non-zero feature values per sample. The $SVM^{perf}$ algorithm \cite{joachims2006training} for linear SVM scales as $O(s m)$. 
The Pegasos algorithm \cite{shalev2011pegasos} improves the complexity to $\BigPolyOh{s/(\lambda\epsilon)}$,
where $\lambda$, and $\epsilon$ are the regularization parameter of SVM and the error of the solution, respectively. 
However, sublinear sparsity of input samples is a strong assumption, and holds only in some special scenarios, such as Bag-of-Words encoding in natural language processing, but rarely holds for data describing physical phenomena, including biomedical data. 

Beyond the classical realm, instead of the solver \cite{van2019quantum} we used, two alternative quantum solvers based on the interior point approach were also introduced recently, an SDP/LP \cite{kerenidis2018quantum} and a dedicated LP \cite{casares2019quantum} solver.
However, both of the quantum interior point solvers come with computational complexity challenges that make them less appealing for training sparse models. The interior point dedicated LP solver \cite{casares2019quantum} has complexity of $\BigOh{\sqrt{n}L/\epsilon}$, where $L=\BigOh{\log{n}}$ on average but $L=mn$ in the worst case. Crucially, the complexity also depends on the sparsity of the constraint matrix $A$. However, in SVMs, $A$ is composed of elements $y_i x_i^j$, the feature values, which are unlikely to be spare in many applications, including those involving biomedical data.  The SDP/LP solver \cite{kerenidis2018quantum} has cubic dependence on the condition number $\kappa$ of the intermediate solution matrices of the SDP. In the LP variant, $\kappa$ is ratio of largest to smallest feature weight magnitude, which can easily explode in machine learning applications, where some feature weights are null or close to null. Even if the problem is well-conditioned, the interior-point SDP/LP solver has complexity of $\BigPolyOh{m^2}$. 

An alternative to using a quantum LP solver is to use quantum gradient descent. However, most quantum gradient descent approaches only work reliably for small number of steps, with probability of following the gradient path decreasing exponentially with the number of gradient updates. Recently, a quantum descent method circumventing this problem has been proposed \cite{kerenidis2020quantum}, but only for objective functions where the norm of the gradient decreases with each step, such as quadratic functions. Sparse SVM objective function is piece-wise linear and non-differentiable, with possibility of subgradient norm increase when a step from one linear region to another is made.

Recently, construction of  classical algorithms based on quantum machine learning methods using sampling-based data structure has been proposed \cite{tang2019quantum}, and has led to a dequantization of LS SVM and other machine learning methods under low-rank assumption \cite{chia2019sampling}. A dequantized SDP solver has also been proposed \cite{chia2019quantum} for SDP problems with low rank input matrices. While a linear program $\mathrm{LP}(c,A,b)$ can be seen as an SDP with diagonal matrices formed by $c$ and rows of $A$, for the LP resulting from Sparse SVM these diagonal matrices are full-rank. This suggests that the proposed quantum Sparse SVM cannot be dequantized using existing, low rank-based approaches.

\section*{Acknowledgments}
T.A. is supported by NSF grant IIS-1453658.

\bibliographystyle{unsrt}
\newcommand{\noopsort}[1]{}

\end{document}